\documentclass[10pt,twocolumn,letterpaper,capitalise,noabbrev,nameinlink,capitalize]{article}

\usepackage{cvpr}

\usepackage{glossaries}
\usepackage{amsthm}
\usepackage{graphicx}
\graphicspath{{figures/}}
\usepackage{float}
\usepackage{tabularx}
\usepackage{amsmath,amssymb,amsfonts,nicefrac}
\usepackage[linesnumbered,ruled,vlined]{algorithm2e}

\SetAlFnt{\small}
\SetAlCapFnt{\small}
\SetAlCapNameFnt{\small}
\SetAlCapHSkip{0pt}

\usepackage[pagebackref]{hyperref}
\usepackage[capitalise,capitalize,noabbrev,nameinlink]{cleveref}
\usepackage{acronym}

\usepackage{enumitem}
\usepackage{balance}
\usepackage{xspace}
\usepackage{setspace}
\usepackage{wrapfig}

\usepackage{caption}
\usepackage{subcaption}
\usepackage[dvipsnames,svgnames,x11names,table]{xcolor}
\usepackage{booktabs,colortbl,multirow,array,makecell,tabularray}
\usepackage{overpic}
\usepackage[misc]{ifsym}
\usepackage{algorithmicx}
\usepackage{algpseudocode}

\usepackage{silence}
\makeatletter
\DeclareRobustCommand\onedot{\futurelet\@let@token\@onedot}
\def\@onedot{\ifx\@let@token.\else.\null\fi\xspace}
\def\eg{\emph{e.g}\onedot}

\def\ie{\emph{i.e}\onedot}

\makeatother

\acrodef{cvae}[cVAE]{conditional Variational Auto-Encoder}
\acrodef{icp}[ICP]{Iterative Closest Points}
\acrodef{mocap}[MoCap]{motion-captured}
\acrodef{vit}[ViT]{Vision Transformer}
\acrodef{fid}[FID]{Fréchet Inception Distance}
\acrodef{gs}[GS]{Gaussian Splatting}
\acrodef{3dgs}[3DGS]{3D Gaussian Splatting}
\acrodef{nerf}[NeRF]{Neural Radiance Fields}
\acrodef{mpm}[MPM]{Material Point Method}
\acrodef{fem}[FEM]{Finite Element Method}
\acrodef{sph}[SPH]{Smoothed Particle Hydrodynamics}
\acrodef{flip}[FLIP]{Fluid-Implicit-Particle}
\acrodef{pic}[PIC]{Particle-in-Cell}
\acrodef{fp}[FP]{FLIP/PIC Blend}
\acrodef{xfem}[XFEM]{Extended Finite Element Method}
\acrodef{bem}[BEM]{Boundary Element Method}
\acrodef{sds}[SDS]{Score Distillation Sampling}
\acrodef{sh}[SH]{Spherical Harmonic}
\acrodef{dssim}[D-SSIM]{Structural Dissimilarity Index Measure}
\acrodef{cdm}[CDM]{Continuum Damage Mechanics}
\acrodef{cdmpm}[CD-MPM]{Continuum Damage Material Point Method}
\acrodef{pff}[PFF]{Phase-Field Fracture}
\acrodef{nacc}[NACC]{Non-Associated Cam-Clay}
\acrodef{fm}[FM]{Fracture Mechanics}
\acrodef{pca}[PCA]{Principal Component Analysis}
\acrodef{pbr}[PBR]{Physically Based Rendering}

\theoremstyle{definition}

\newcommand{\modelname}{\textbf{GaussianFluent}\xspace}

\newcommand{\phong}{Blinn-Phong reflection model\xspace}

\newcommand{\supp}{\textit{Supplementary}\xspace}

\makeatletter
\renewcommand{\paragraph}{%
  \@startsection{paragraph}{4}{\z@}%
  {1ex plus 0.5ex minus 0.2ex} 
  {-1em}                      
  {\normalfont\normalsize\bfseries} 
}
\makeatother

\usepackage{amsmath,amsfonts,bm}

\def\eqref#1{equation~\ref{#1}}

\def\1{\bm{1}}

\DeclareMathAlphabet{\mathsfit}{\encodingdefault}{\sfdefault}{m}{sl}
\SetMathAlphabet{\mathsfit}{bold}{\encodingdefault}{\sfdefault}{bx}{n}

\def\paperID{}
\def\confName{CVPR}
\def\confYear{2026}

\title{\modelname: Gaussian Simulation for Dynamic Scenes with
Mixed Materials}

\author{
    Bei Huang\textsuperscript{1,2*}, 
    Yixin Chen\textsuperscript{2*$\dagger$}, 
    Ruijie Lu\textsuperscript{1,2},
    Gang Zeng\textsuperscript{1},
    Hongbin Zha\textsuperscript{1},
    Yuru Pei\textsuperscript{1$\dagger$},
    Siyuan Huang\textsuperscript{2$\dagger$} \\
    \small \textsuperscript{$\star{}$} Equal Contribution \,\textsuperscript{$\dagger$} Corresponding Authors\\
    \small\textsuperscript{1}State Key Laboratory of General Artificial Intelligence, Peking University\\
    \small\textsuperscript{2}State Key Laboratory of General Artificial Intelligence, BIGAI\\
    \url{https://hb-pencil-zero.github.io/GaussianFluent/}
}

\begin{document}

\twocolumn[{%
\renewcommand\twocolumn[1][]{#1}%
\maketitle
\begin{center}
    \centering
    \captionsetup{type=figure}
    \includegraphics[width=\textwidth]{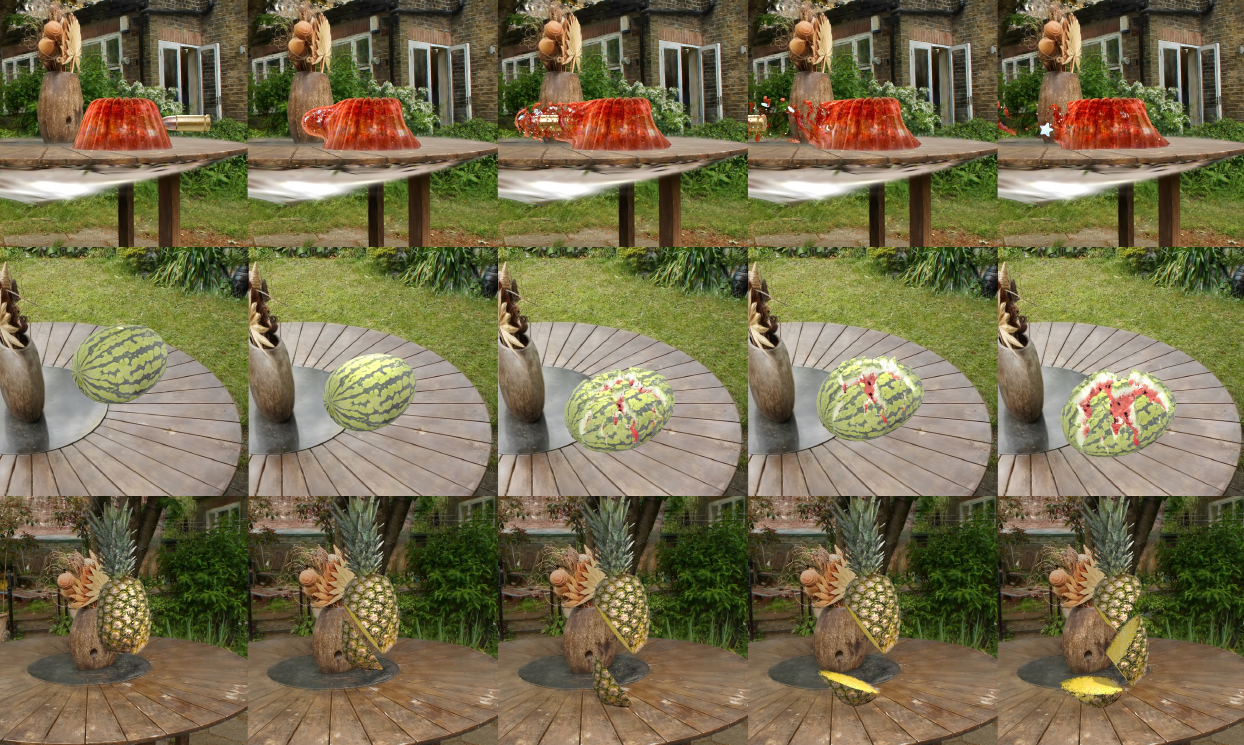}
  \captionof{figure}{\textbf{Physical simulation of dynamic object states with \acl{3dgs}.} \modelname is capable of generating realistic internal texture, simulating and rendering complex object dynamics (\eg, elastic deformation, fracture, and slicing) with mixed materials (\eg, jelly with internal blue sugar penetrated by a rigid bullet in top row), in response to different lighting conditions.}
    \label{fig:teaser}
\end{center}%
}]

\begin{abstract}
\ac{3dgs} has emerged as a prominent 3D representation for high-fidelity and real-time rendering. Prior work has coupled physics simulation with Gaussians, but predominantly targets soft, deformable materials, leaving brittle fracture largely unresolved. This stems from two key obstacles: the lack of volumetric interiors with  coherent textures in \acs{gs} representation, and the absence of fracture-aware simulation methods for Gaussians. To address these challenges, we introduce \modelname, a unified framework for realistic simulation and rendering of dynamic object states. First, it synthesizes photorealistic interiors by densifying internal Gaussians guided by generative models. Second, it integrates an optimized \ac{cdmpm} to enable brittle fracture simulation at remarkably high speed. Our approach handles complex scenarios including mixed-material objects and multi-stage fracture propagation, achieving results infeasible with previous methods. Experiments clearly demonstrate \modelname's capability for photo-realistic, real-time rendering with structurally consistent interiors, highlighting its potential for downstream application, such as VR and Robotics.

\end{abstract}

\section{Introduction}

\acf{3dgs}~\citep{kerbl3Dgaussians} has recently emerged as a prominent and highly effective technique for high-fidelity, real-time rendering of complex 3D scenes, achieving state-of-the-art rendering quality with exceptional efficiency. Despite its remarkable success, modeling dynamic scenes within the \ac{gs} framework, especially the physics simulation of consistent evolution of multi-material objects, still presents significant challenges. This difficulty stems from two primary fundamental issues.

First, as a surface-based method, \acl{gs} inherently lacks representation of internal structures. Consequently, the stress, inertia, and contact-force computations required for physically accurate solid-object simulation remain undefined. More critically, \ac{gs} cannot realistically render the newly exposed surfaces during fracture. For instance, simulating a watermelon falling and fracturing would require modeling its red flesh and black seeds beneath the green rind. However, current \ac{gs} reconstruction methods leave such interiors hollow and textureless, making realistic fracture visualization impossible.

Second, previous \ac{gs} simulation methods, such as PhysGaussian~\citep{xie2024physgaussian}, have largely targeted elastic material dynamics. While subsequent works, such as OmniPhysGS~\citep{lin2025omniphysgs} and Pixie~\citep{le2025pixiefastgeneralizablesupervised}, have automated the estimation of material parameters (e.g., Young's modulus and Poisson's ratio), they remain confined to elastic deformation and do not extend simulation capabilities to more complex materials. Consequently, methods capable of simulating brittle fracture and topological changes within the \ac{gs} framework are still absent.
Existing point-cloud fracture methods~\citep{wolper2019cd} are incompatible with \ac{gs} representation: they lack a continuous return-mapping scheme, resulting in physically implausible fracture dynamics, and their reliance on CPU-bound execution, with limited parallelism, imposes severe performance bottlenecks. These limitations hinder the application of \ac{gs} to realistic scenarios involving complex fracture behaviors and dynamic structural changes.


To address these challenges, we introduce \modelname, a novel framework to populate \ac{gs} interiors and simulate complex object dynamics such as brittle fracture and bullet impacts, with physically accurate responses under dynamic lighting. More specifically, we introduce:
1) \textbf{internal texture synthesis}, a novel pipeline that synthesizes realistic and consistent internal structures and textures for GS by leveraging publicly available generative models, requiring no additional training data,
and 2) \textbf{optimized CD-MPM for \ac{gs}}, where we augment the current \ac{gs} simulation framework with an optimized integration of \ac{cdmpm}, resolving instability issues in the previous algorithm and implementing GPU parallelism. This enables physics-plausible brittle fracture simulation with substantial real-time rendering. 

We validate \modelname on a suite of challenging scenarios involving food, liquids, and fruits, where internal and external appearances differ significantly, and materials span brittle solids, viscoelastic gels, and soft tissues. Our experiments cover diverse topological changes, including dynamic fracturing, elastoplastic deformation, slicing, and high-velocity bullet impacts, such as a bullet shooting through jelly and crossing over it, a watermelon falling down onto a table and fracturing, and milk falling onto a table with splashing. Results show that our method effectively reconstructs structurally coherent internal GS primitives with realistic textures and achieves high-fidelity simulation and rendering of dynamic scenes, substantially outperforming existing methods.

\section{Related work}
\subsection{Deformation-Predicted Dynamic Scenes}
\ac{nerf}~\citep{mildenhall2021nerf,muller2022instant,barron2021mip,barron2023zip,barron2022mip,chen2022tensorf} and \ac{3dgs}~\citep{kerbl3Dgaussians, yu2024mip, huang20242d, chen2024survey} have recently emerged as two prominent approaches for scene reconstruction, largely due to their ability to produce photo-realistic and efficient renderings. However, both methods primarily focus on static scenes and lack inherent support for modeling dynamic environments. To address this limitation, subsequent works incorporate deformation fields into neural radiance fields~\citep{pumarola2021d, park2021hypernerf, tretschk2021non} and Gaussian primitives~\citep{wu20244d, yang2024deformable, huang2024sc, wan2024superpoint, liang2024feed, luiten2024dynamic} to capture scene dynamics. Despite these advancements, existing approaches are typically limited to replaying observed motion trajectories rather than enabling further simulation or interaction, thereby restricting their generalization capability. Moreover, the modeling of motion in deformable Gaussians often lacks physically grounded constraints: each Gaussian is assigned an independent deformation vector without regard to physical plausibility, which can result in unrealistic or implausible dynamics.

\subsection{Physics-Simulated Dynamic Scenes for GS}
\acf{gs} is inherently compatible with the \ac{mpm} physics simulation framework, as its representation is composed of particle-like primitives, which provide a unified explicit foundation for both simulation and rendering. PhysGaussian~\citep{xie2024physgaussian} pioneers this direction by associating physical properties with Gaussian primitives and employing the \ac{mpm} for physically based simulation. Subsequent works~\citep{huang2024dreamphysics, zhang2024physdreamer, liu2024physics3d} extend this framework by either learning physical properties from generative priors~\citep{blattmann2023stable, xing2024dynamicrafter, wang2023modelscope, lin2025omniphysgs}, enabling automated physical parameter optimization. 
However, these methods still cannot model highly dynamic scenes, primarily due to the absence of simulation models suitable for brittle fracture. 
Furthermore, existing methods generally neglect the plausibility of internal textures that become visible when objects tear or break. 
FruitNinja~\citep{wu2025fruitninja} addresses internal texture generation for static \ac{gs} reconstructions of fruits using a diffusion model fine-tuned on a self-collected dataset, which is extremely costly and lacks generalizability.

Building upon Continuum Damage Mechanics (CDM) used in point cloud~\citep{simo1987strain, matsuoka1999cam, bourdin2000numerical}, we develop an optimized CD-MPM formulation for \ac{3dgs} that delivers realistic brittle fracture on mixed-material objects, and pair it with an efficient internal texture filling pipeline.

\section{Method}

\label{sec:method}
We propose \modelname to enable realistic simulation of dynamic scenes, particularly material fracture, within the \ac{3dgs} framework. The overall framework is shown in \cref{fig:workflow}. Our method first generates internal structures and textures for \ac{gs} representations, followed by simulating fracture dynamics using an optimized CD-MPM framework to achieve diverse simulations across solid objects to fluids.

\subsection{Internal Filling for 3D Gaussian Splatting}
\label{sec:gs_inpaint}

\subsubsection{Internal Volume Initialization}
\label{sec:internal_fill}

\begin{figure*}[ht!]
    \centering
    \includegraphics[width=\textwidth]{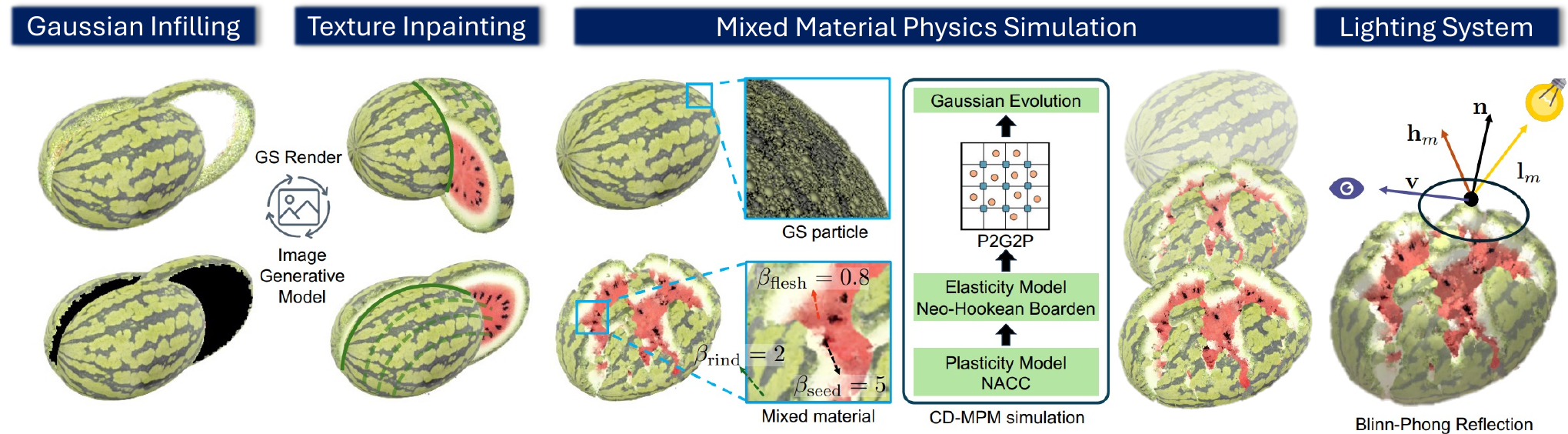}
    \caption{\textbf{Overview of \modelname}. Our model first populates Gaussians in the internal volume and generates interior realistic texture with pretrained image generative models (\cref{sec:gs_inpaint}). We then incorporate optimized CD-MPM simulation with mixed materials for Gaussian Splatting (\cref{sec:cdmpm}) and introduce Blinn-Phong reflection in the rendering pipeline ( \supp \cref{sec:phong}).}
    \label{fig:workflow}
\end{figure*}

Standard \ac{3dgs} primarily captures external surfaces, leaving interiors undefined, which is problematic for simulating interactions like cutting that expose internal structures. Our method first populates the interior volume and then textures it, as illustrated in \cref{fig:workflow}. 

To initialize the internal volume, we first train an initial \ac{3dgs} model of the target object from multiview images. To prevent large Gaussians from straddling boundaries and ensure a clear exterior-interior separation~\citep{liu2024atomgs,wu2025fruitninja}, we augment the standard rendering loss with a scale regularization:

\begin{equation}\label{eq:scale_reg_condensed}
\mathcal{L}_{\text{total}} = \, \mathcal{L}_{\text{MSE}} + \, \mathcal{L}_{\text{SSIM}} + \lambda \sum_{i=1}^{N} \|\mathbf{s}_i\|_2^2,
\end{equation}
where $\mathbf{s}_i$ are the scale parameters of Gaussian $i$, and $\lambda$ controls regularization strength. This encourages smaller, more localized Gaussians, crucial for interior definition and plausible performance under relighting.

\begin{figure}[t!]
    \includegraphics[width=\linewidth]{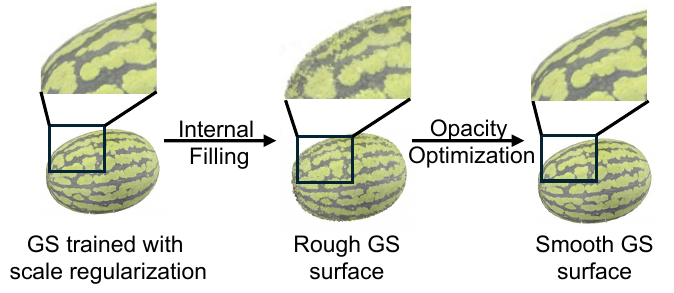}
    \caption{\textbf{Internal Gaussian filling and refinement.} The opacity optimization improves the smoothness of the \ac{gs} surface after internal filling, beneficial for texture inpainting and simulation.}
    \label{fig:fillstep1}
\end{figure}

Next, we identify the object boundary as the high-density regions. Given a resolution $n$, we uniformly discretize the scene space into $n^3$ grids and compute the density field $d(x)$ for each grid center by accumulating contributions from its neighboring Gaussians $P$:
\begin{equation}
\label{eq:opacity_field_condensed}
d(x)=\sum_{p\in P}\alpha_p \exp\!\Big(-\tfrac{1}{2}(x-x_p)^{T}\mathbf{A}_p^{-1}(x-x_p)\Big),
\end{equation}

where \(\alpha_p\), \(x_p\), and \(\mathbf{A}_p\) denote the opacity, \ac{gs} center, and covariance of Gaussian \(p\), respectively. Grids with \(d(x)\ge \tau_d\) are marked high-density;  these high-density grids are extracted as the object boundary. New internal Gaussians are then initialized inside the enclosed volume, following prior practice in PhysGaussian \cite{xie2024physgaussian}.

This initial density-based filling can be imprecise, potentially creating Gaussians outside the true boundary due to sensitivities to surface geometry and threshold choice~\citep{wu2025fruitninja}. To refine this, we perform an opacity-only optimization for all new internal Gaussians using the rendering loss by fixing other attributes. This drives the opacity of extraneous Gaussians to zero. Finally, we prune Gaussians with opacity near zero, resulting in a clean, well-defined solid volume representation suitable for subsequent texturing and simulation, as shown in \cref{fig:fillstep1}.

\subsubsection{Internal Texture Generation}
\label{texture_generation_detailed_condensed}

Once the interior volume is populated, assigning plausible internal textures is the next crucial step. Generating multi-view and spatially coherent internal textures is a particularly significant challenge due to scarce training data for object interiors~\citep{poole2022dreamfusion, liu2024one}.
Thus, we propose a training-free two-stage approach: an initial texture generation via single-view inpainting, followed by iterative multi-axis refinement, as illustrated in \cref{fig:workflow}.

\paragraph{Coarse Texture Initialization}
We first establish a coarse internal texture by uniformly slicing the object into  slices along the X-axis and inpainting each slice from its frontal viewpoint. For each slice, we render its initial appearance $\mathbf{C}_{\text{initial}}$ and an internal region mask $\mathbf{M}_{\text{init}}$. The masked region in $\mathbf{C}_{\text{initial}}$ is then inpainted using a generative model, \eg, MVInpainter~\citep{cao2024mvinpainter}, which is guided by a text prompt $\mathcal{P}$ and a reference image generated by Stable Diffusion XL (SD-XL) to produce the consistent target image $\mathbf{C}_{\text{inpaint}}$. Each internal Gaussian $i$ whose 2D projection $\mathbf{u}_i$ falls within the inpainted region then samples its color $\mathbf{c}_i$ from $\mathbf{C}_{\text{inpaint}}$ using bilinear interpolation. Its zeroth-order spherical harmonic (SH) coefficient, $\mathbf{sh}^0_i$, is initialized as:
\begin{equation}
\label{eq:sh0_init_detailed_condensed}
\mathbf{sh}^0_i = \frac{\mathbf{c}_i - 0.5}{C_0},
\end{equation}
where the constant $C_0 = 1 / (2\sqrt{\pi})$. Higher-order SH coefficients for these internal Gaussians are simply initialized to zero, ensuring an initially isotropic appearance derived from the inpainted texture.

\paragraph{Iterative Texture Refinement}
The single-view initialization, while providing a reasonable start, still lacks consistency across the 3D internal structure and different viewing directions. To achieve  consistency, we therefore iteratively refine the texture across all three primary axes (X, Y, and Z). This refinement is carefully guided by text-prompted image inpainting using image generative models like SD-XL~\citep{podell2023sdxlimprovinglatentdiffusion}; detailed prompts are provided in \supp\cref{appx:details}.

Inspired by the iterative corrective philosophy of SDS~\citep{poole2022dreamfusion}, we perform successive low-strength inpainting updates.
The core refinement loop, repeated per iteration, consists of two main steps:
1) \textit{Generative Inpainting of Slices}: We uniformly space slices along each of the X, Y, and Z axes. For every slice, we render its axis-aligned orthographic view and an internal structure mask; these inputs are passed to SD-XL with a low inpaint strength, constraining denoising so that edits incrementally inject new internal details while maintaining the global structure.
2) \textit{Gaussian Optimization}: The newly inpainted 2D images from all  slices serve as optimization targets. The SH coefficients of the internal \ac{gs} are optimized for small steps to minimize the rendering discrepancy against these inpainted images.

This two-step cycle is systematically repeated until the optimization loss converges or a maximum number of iterations is reached, ultimately yielding an internally consistent and highly detailed 3D texture. Our successive low-strength inpainting strategy produces sharp and realistic textures, in stark contrast to vanilla SDS, which leads to blurry and oversaturated results~\citep{wang2023prolificdreamer,alldieck2024score,lukoianov2024score}. Since the consistent images generated by MVInpainter and the internal slices are inherently co-dependent through their intersections on orthogonal views, the iterative refinement drives the optimization toward tri-axial consistency. For example, given a three-layer cake as the reference image, MVInpainter ensures all X-slices exhibit the three-layer structure; subsequently, the low-strength inpainting by SD-XL propagates this constraint to the Y- and Z-slices, as any deviation would contradict the established X-slice structure.

\subsection{CD-MPM in GS with Mixed Materials}
\label{sec:cdmpm}
We extend the \ac{3dgs} simulation framework by incorporating the CD-MPM with support for mixed materials. Similar to PhysGaussian~\citep{xie2024physgaussian}, each 3D Gaussian primitive in our framework is assigned physical properties, including mass, velocity, volume, and stress, and interacts with other particles via a background Eulerian grid. Our GPU parallelization implementation for efficient physical simulation is detailed in \supp \cref{appx:gpu_parallel}.

\paragraph{Initialization}
We initialize covariances only for newly added interior Gaussians, assigning each a spherical covariance whose radius directly corresponds to its per-particle volume, \ie, the cell volume divided by the number of particles in the cell. This initialization strategy ensures strong spatial consistency between the Gaussian representation and the MPM discretization. The physical material parameters of the Gaussians, such as Young's modulus, Poisson's ratio, friction angle, mass density, fracture control parameters $\beta$ and $\alpha$, etc., are manually defined following PhysGaussian and CD-MPM.

\paragraph{\ac{gs} Property Evolution with MPM}
Let \( \mathbf{X} \) denote the reference \ac{gs} state before simulation, and \( \mathbf{x} \) the state after simulation. Continuum mechanics describes motion via a time-dependent deformation map as follows:
\begin{equation}
\mathbf{x} = \boldsymbol{\varphi}(\mathbf{X}, t). 
\end{equation}
Here, \( \boldsymbol{\varphi} \) represents the MPM simulation function. The deformation gradient \( \mathbf{F}_p(t) \) is defined as
\begin{equation}
\mathbf{F}_p(t) = \frac{\partial \mathbf{x}}{\partial \mathbf{X}} = \frac{\partial \boldsymbol{\varphi}(\mathbf{X}, t)}{\partial \mathbf{X}},
\end{equation}
which encodes both local rigid deformation (rotation) and non-rigid deformation (stretch and shear). For each simulation step, we apply \( \mathbf{F}_p(t) \) to the \ac{gs}'s covariance and spherical harmonics to achieve physics-plausible simulation results. For more details, please refer to \supp \cref{appx:mpm}.

\paragraph{Fracture Mechanism}

\begin{figure}[t!]
    \includegraphics[width=\linewidth]{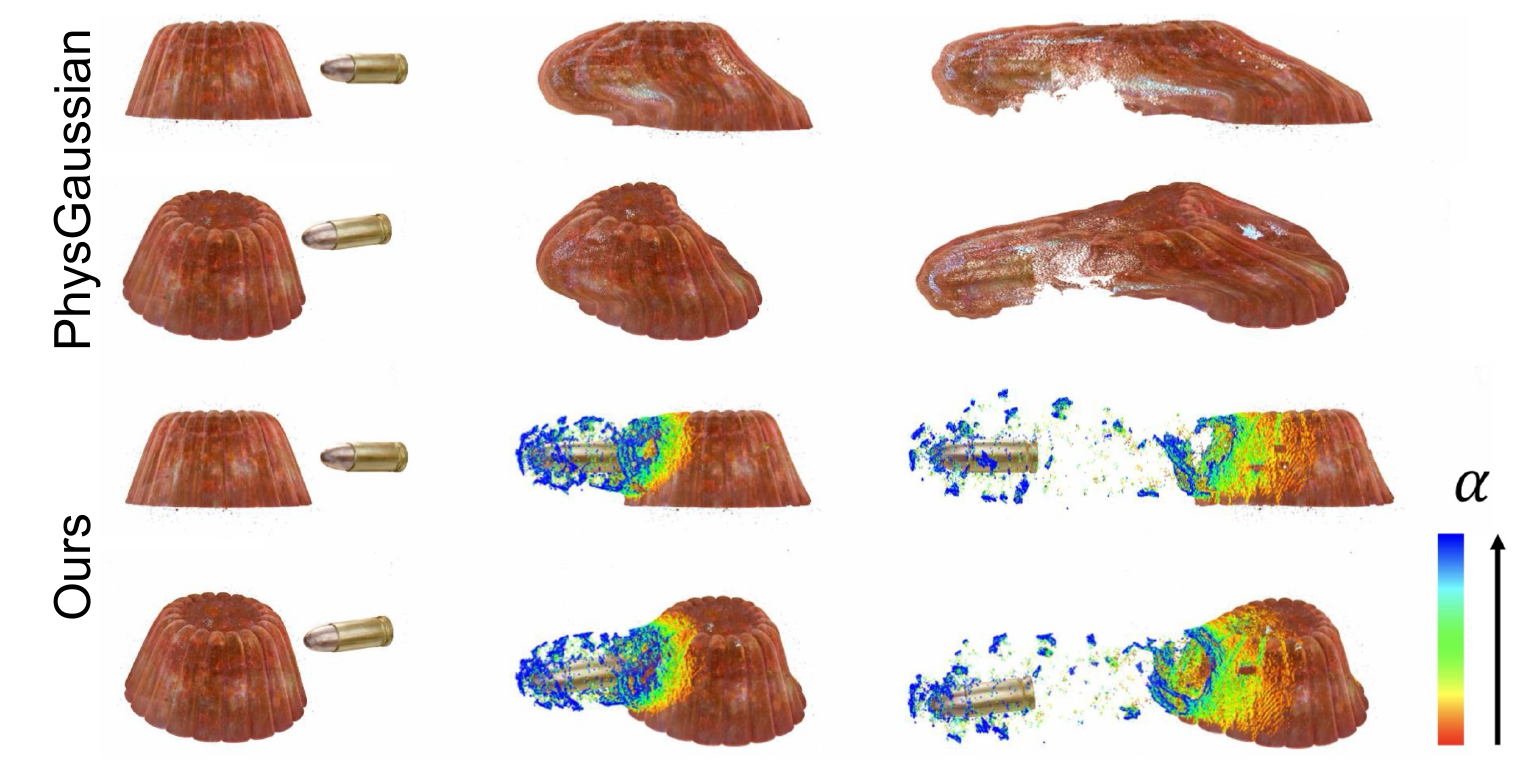}
\caption{\textbf{A jelly-like material is shot with a bullet.} We compare our method with PhysGaussian to demonstrate the effectiveness of our simulation and visualize the damage variable \(\alpha\).}
    \label{fig:heatmap}
\end{figure}

We model brittle fracture by tracking the deformation \( \mathbf{F}_p(t) \) of each \ac{gs}. A softening law reduces its stress-generating capacity with increasing deformation. Fracture is not triggered by a sharp threshold but emerges when this capacity becomes negligible and fails to sustain internal forces. We decompose \( \mathbf{F}_p(t) \) into rigid and non-rigid components; only the latter, comprising volumetric stretch \(p\) and shear distortion \(q\), contributes to fracture. A square under volumetric stretch becomes a scaled orthogonal rectangle, whereas pure shear turns it into an area-preserving parallelogram with skewed angles. The elastic, stress-generating region is defined by a yield surface \( y(p,q) \le 0 \). With accumulating deformation, this surface contracts in the \( (p,q) \)-plane, diminishing the sustainable elastic stress. Fracture occurs as this residual capacity vanishes. The \ac{nacc} model specifies this surface via the equation \( y(p,q; p_0, \beta, M)=0 \). More specifically,
\begin{align*}
y(p, q; p_0, \beta, M)
  &= q^2(1 + 2\beta) + M^2(p + \beta p_0)(p - p_0),\\[6pt]
p_0 &= K \sinh(\xi \max(-\alpha, 0)).
\end{align*}
$\beta, M, K, \xi$ are all predefined hyperparameters, following the setting of CD-MPM, $p$ is the volumetric stretch magnitude, and $q$ is the shear magnitude. $\alpha$ is the key damage variable. At each step, we apply return mapping to enforce  $y \le 0$ and update $\alpha$ to evolve the yield surface $y$. 

\paragraph{Continuous Return Mapping}

At each step, a trial state \( (p^{\mathrm{tr}}, q^{\mathrm{tr}}) \) is formed and evaluated by the yield function \( y^{\mathrm{tr}} = y(p^{\mathrm{tr}}, q^{\mathrm{tr}}) \). Only the region where \( y \leq 0 \) is physically meaningful. Therefore, when \( y > 0 \), it is necessary to project \( (p^{\mathrm{tr}}, q^{\mathrm{tr}}) \) onto the ellipsoid such that \( y = 0 \). In CD-MPM, this projection involves two possible cases:

\begin{figure*}
    \includegraphics[width=\textwidth]{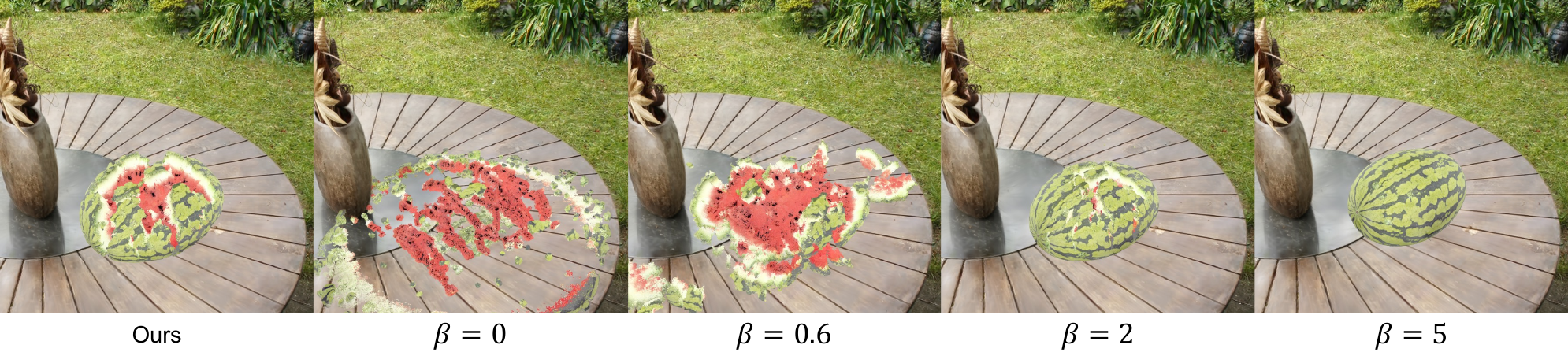}
    \caption{\textbf{Comparison between our mixed material modeling and fixed $\beta$ setting.} Our approach assigns distinct $\beta$ values, \ie, 2, 0.6, and 5, to the rind, flesh, and seed, respectively. This yields more realistic simulation results compared to settings that apply a single, uniform $\beta$ value to the entire watermelon. }
    \label{fig:beta}
\end{figure*}

\begin{enumerate}[leftmargin=*]
\item Exterior pressures (\( p^{\text{tr}} \ge p_0 \) or \( p^{\text{tr}} \le - \beta p_0 \)): tip projection, where \( p^{\text{tr}} \ge p_0 \Rightarrow (p_0,0) \), and \( p^{\text{tr}} \le - \beta p_0 \Rightarrow (-\beta p_0,0) \).
\item Interior pressures (\( -\beta p_0 < p^{\text{tr}} < p_0 \)): connect \( (p^{\text{tr}}, q^{\text{tr}}) \) to \( (p_c,0) \) with the ellipse center \( p_c = \frac{-\beta p_0 + p_0}{2} = \frac{1-\beta}{2} p_0 \), where the intersection with the yield ellipse gives \( (p_{\text{new}}, q_{\text{new}}) \).
\end{enumerate}

The connection to the fixed center \( (p_c,0) \) causes return-map discontinuities at \( p=p_0 \) and \( p=-\beta p_0 \). At the right boundary $p$, letting \( q^{\text{tr}} \to \infty \) shows a jump:
\begin{align*}
\lim_{\varepsilon \to 0^+} \lim_{q^{\text{tr}} \to \infty} R(p_0 - \varepsilon, q^{\text{tr}}) &= \left(\frac{1-\beta}{2}p_0, \frac{M(\beta+1)}{2\sqrt{1+2\beta}}p_0\right), \\
\lim_{\varepsilon \to 0^+} \lim_{q^{\text{tr}} \to \infty} R(p_0 + \varepsilon, q^{\text{tr}}) &= (p_0, 0).
\end{align*}

We present a schematic diagram in \supp ~\cref{fig:return_mapping} to illustrate the discontinuity jump problem of this projection. Specifically, approaching \( p \to p_0^{-} \) with \( q^{\text{tr}} \to \infty \) maps the trial state to the upper apex of the yield ellipse, whereas \( p \to p_0^{+} \) maps it to the right tip \( (p_0,0) \).
This jump triggers numerical instability: a machine-precision fluctuation \( \delta \) about \( p_0 \) can map an identical geometric state to completely different return points. To resolve this instability, we regularize the projection by introducing a \textbf{dynamic point}, \( (p_c', 0) \), which smoothly adapts to the \( (p, q) \) and ensures a continuous mapping. We define this new point as:
\begin{equation}
\label{eq:dynamic_point}
p_c' = p_c + \phi_k(p^{\text{tr}})(p^{\text{tr}} - p_c),
\end{equation}
where \(\phi_k(p^{\text{tr}}) = \left| \frac{p^{\text{tr}} - p_c}{p_0 - p_c} \right|^k\) and \( p_0 - p_c \) is the semi-major axis of the ellipse.

This modified scheme can be regarded as an extension of the original approach, replacing the fixed point \( (p_c, 0) \) with a dynamic point \( (p_c', 0) \). For any finite \( k \), we have \( \lim_{\varepsilon \to 0^+} p_c'(p_0 - \varepsilon) = p_0 \), indicating continuity:
\begin{align*}
\lim_{\varepsilon \to 0^+} \lim_{q^{\text{tr}} \to \infty} R(p_0 - \varepsilon, q^{\text{tr}}) &= (p_0,0), \notag \\
\lim_{\varepsilon \to 0^+} \lim_{q^{\text{tr}} \to \infty} R(p_0 + \varepsilon, q^{\text{tr}}) &= (p_0,0). \notag
\end{align*}

Moreover, it recovers the original discontinuous scheme in the limit as \( k \to \infty \), because for any \( p_{\mathrm{tr}} \) in \( (-\beta p_0, p_0) \), we have
\( \lim_{k \to \infty} p_c' = \lim_{k \to \infty} (p_c + \left| \frac{p^{\mathrm{tr}} - p_c}{p_0 - p_c} \right|^k) =p_c \).
In our practical  implementation, we choose \( k=2 \), as it provides a robust and smooth projection. After projection, we compose \( p \) and \( q \) to obtain the \( \mathbf{F}_p \). We define \( J_{\mathrm{tr}} = \det \mathbf{F}_p^{\mathrm{tr}} \) and \( J_{\mathrm{new}} = \det \mathbf{F}_p^{\mathrm{new}} \), and update the hardening parameter via
\( \alpha \leftarrow \alpha + \ln \left( \frac{J_{\mathrm{tr}}}{J_{\mathrm{new}}} \right) \).
We present an example of \( \alpha \) heatmap in ~\cref{fig:heatmap} to visualize the changes in \( \alpha \) within the jelly. As the \( \alpha \) value progressively increases, the corresponding regions of the jelly undergo fracturing. Further details are provided in \supp ~\cref{sec:cdmpm_supp}.

\paragraph{Mixed material simulation}
Unlike PhysGaussian, which assumes uniform material properties, our method supports more realistic and complex simulations by assigning different \( \beta \) to various parts of an object, such as the seed, flesh, and rind of a watermelon. This requires segmenting both external and internal structures through existing segmentation methods~\citep{yang2024sampart3d,liu2025partfield}, part-aware object generation~\citep{yang2025omnipart,zhang2025bang}, or heuristics.
For example, to realistically model a watermelon fracture, we assign $\beta$ values based on the color of the \ac{gs}, \ie, a high $\beta$ to the black seeds, a low $\beta$ to the red flesh, and a middle $\beta$ to the green rind, where the value range of $\beta$ follows the design in CD-MPM.
As shown in \cref{fig:beta}, our mixed material approach produces more realistic results, whereas using a uniform material leads to visual artifacts and unnatural fracture patterns. The lollipop shattering scene in \cref{fig:simulation_comparison} further demonstrates the cracks that PhysGaussian cannot generate.

\paragraph{Extension to Fluid Simulation}
Beyond brittle fracture, our framework is also applicable to simulating fluid-like materials. Many real-world
scenarios involve both solid and fluid phases, such as milk flowing
on a table. Rather than coupling separate solid and fluid solvers,
we achieve unified simulation by adjusting material parameters within
the same CD-MPM framework.

The key insight is that our \ac{nacc} model naturally degenerates to
fluid behavior when \( p_0 \to 0 \). In this limit, the elastic region vanishes, and the material loses its resistance to separation, allowing it to flow freely like a liquid. Combined with
appropriate elastic parameters, \eg, Young's modulus \( E \) and Poisson's ratio \( \nu \), that characterize fluid-like compressibility, this
configuration produces physically realistic fluid behavior. We exploit this property to simulate fluids while preserving the continuous return mapping scheme, enabling a unified simulation framework that handles both solids and fluids without separate treatment. In practice, we set \( \alpha_0 \) and \( \beta \) close to zero to achieve \( p_0 \to 0 \).

\begin{figure*}[ht!]
\centering
\includegraphics[width=\textwidth]{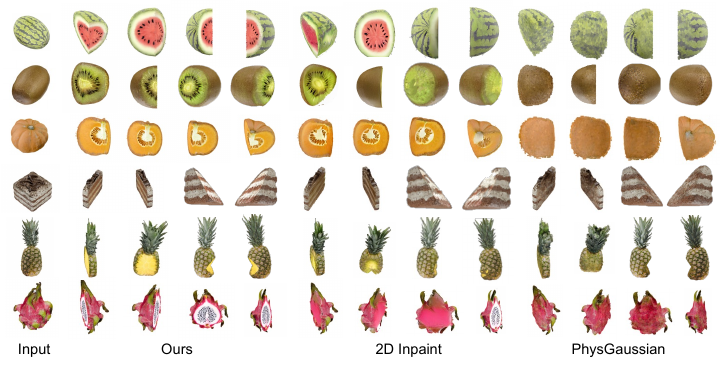}
\caption{\textbf{Qualitative comparison of internal texture filling.} Our method yields more realistic and consistent interior textures from \ac{gs} rendering. PhysGaussian copies exterior colors to internal Gaussians, resulting in blurring, while 2D inpainting fails on oblique views and suffers from multi-view inconsistency.}
\label{fig:interior_comparison}
\end{figure*}

Remarkably, this fluid-like behavior is \textit{independent of the
underlying elastic constitutive model}. When \( p_0 \) is sufficiently
small, our NACC model makes the elastic phase negligible. Thus, materials with vastly different elastic properties, whether using linear elasticity or
hyperelastic models, such as Neo-Hookean for rubber-like materials, all
exhibit identical fluid behavior under the same low-\( p_0 \) regime.
Similarly, retaining a small but non-zero yield surface produces
clay-like viscoplastic flow.

\cref{fig:example} in \supp showcases simulations of sandcastles and milk
as extensive examples. The milk flows down the table naturally, and the sandcastle breaks down under gravity   with the
Neo-Hookean elastic constitutive model, exhibiting  fluid and viscous behavior through plastic dissipation. This demonstrates that our framework
spans the various spectra, from brittle solids to flowing fluids,
all within a single, unified model.

\section{Experiment}
We conduct experiments on both internal texture filling and the physical simulation of dynamic scenes. 
For a more intuitive visualization of our results, we refer the reader to the \textbf{\textit{supplementary videos}}.

\subsection{Internal Texture Filling}

\begin{table}[htbp]
\small
\centering
\caption{\textbf{Quantitative internal filling comparison.} PhysGaussian's direct color copying results in blurred textures, whereas 2D inpainting fails on oblique viewpoints. }
\label{tab:interior_clip}
\begin{tabular}{lcc}
\toprule
Method & CLIP Score $\uparrow$ & User study $\uparrow$ \\
\midrule
PhysGaussian & 22.3 & 3.57\% (3/84) \\
2D Inpainting & 30.1 & 25.00\% (21/84) \\
Ours & \textbf{35.4} & \textbf{71.43\% (60/84)} \\
\bottomrule
\end{tabular}
\end{table}

We evaluate the quality of the generated interior texture, both quantitatively and qualitatively, against PhysGaussian and 2D Inpainting.
We report CLIP scores~\citep{radford2021learning} and conduct a user study, where participants are asked to select the best internal filling results. As presented in~\cref{tab:interior_clip}, our method achieves the highest CLIP score, significantly outperforming PhysGaussian and 2D inpainting. These results indicate that the interior textures generated by our approach exhibit superior semantic consistency with the target descriptions. Prompt details are shown in \supp ~\cref{appx:details}.

\cref{fig:interior_comparison} provides a qualitative comparison of rendered internal structures, and our method produces highly realistic and visually detailed results. For instance, the figure showcases the distinct seeds and flesh texture within a watermelon, a spherical cross-section of a kiwi that reveals its characteristic patterns, and an oblique slice through a cake displaying its clearly defined layers. These high-fidelity results stand in sharp contrast to those from PhysGaussian, which appear significantly blurrier and less defined. Furthermore, while 2D inpainting can produce plausible individual slices, it fails to maintain 3D consistency across different views, resulting in visually unconvincing volumetric representations. In addition to static textures, our method achieves realistic dynamic rendering during simulation, effectively capturing authentic material behavior under various physical conditions, as detailed in the next section.

\subsection{Physics Simulation for dynamic scenes}
\label{sec:physics_simulation}

\begin{figure*}
    \centering
    \includegraphics[width=\linewidth]{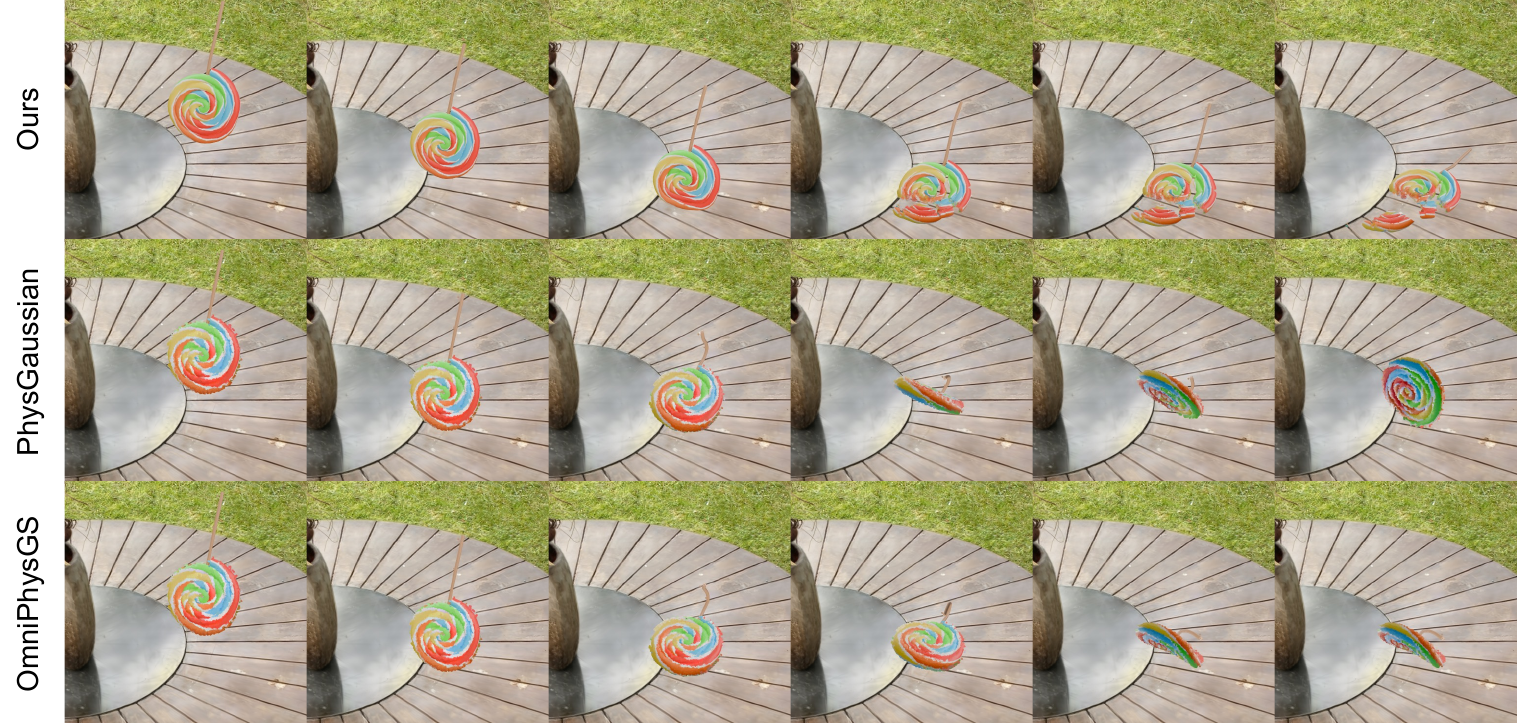}
    \caption{
        \textbf{Qualitative comparison of object state simulation.} 
        We present a comparison for a lollipop, where our result correctly simulates its fracture of mixed materials, outperforming PhysGaussian and OmniPhysGS.
    }
    \label{fig:simulation_comparison}
\end{figure*}
Quantitative evaluations of physics simulations further validate the performance of our method using both CLIP similarity scores and a perceptual user study, where participants are asked to choose the most realistic simulation outcome from our method and the baselines using the same form as \cref{fig:simulation_comparison}. As shown in~\cref{tab:simulation_clip}, our method achieves the highest CLIP similarity score and user preference, substantially outperforming PhysGaussian and OmniPhysGS. The perceptual study demonstrates that our results align with users' understanding of realistic dynamic evolution and conform to intuitive physical commonsense. The higher CLIP score affirms that our simulation outcomes are not only visually more convincing but also semantically more accurate. 

\begin{table}[t]
\small
\centering
\caption{\textbf{Dynamic scene simulation comparison.} Our method significantly outperforms baselines. PhysGaussian fails to produce brittle fracture, and OmniPhysGS is constrained by the PhysGaussian framework.}
\label{tab:simulation_clip}
\begin{tabular}{lcc}
\toprule
Method & CLIP Score $\uparrow$ & User study $\uparrow$ \\
\midrule
PhysGaussian & 12.2 & 3.84\% (1/26) \\
OmniPhysGS & 13.1 & 7.69\% (2/26) \\
Ours & \textbf{22.7} & \textbf{88.46\% (23/26)} \\
\bottomrule
\end{tabular}
\end{table}

\paragraph{\textbf{Diverse Object Simulation}}
We conduct an extensive series of qualitative experiments, as shown in \cref{fig:teaser,fig:simulation_comparison}, and \cref{fig:example,fig:example2} in \supp, to further substantiate the broad applicability and robustness of our framework for various objects with significantly different material properties. The set included elastic materials like jelly, sliceable fruits such as pineapples and kiwis, brittle objects like watermelons, fluids including milk, and granular structures like sandcastles. They provide compelling visual evidence of our model's capabilities under different lighting conditions in diverse scenarios. For example, the top example in \cref{fig:teaser} illustrates the deformation of a jelly when struck by a bullet, highlighting not only its elastic response but also the detailed internal expulsion of rigid sugar.
More examples are included in the supplementary video.

\paragraph{\textbf{Mixed-Material Physics Simulation}}
Our method simulates complex fractures and deformations for objects with different material responses. \mbox{\cref{fig:simulation_comparison}} highlights this with a challenging scenario: a lollipop shatters on impact while its wooden stick remains intact. This ability to model mixed-material physics is visibly more detailed and realistic than prior works. This is also demonstrated in our simulation of a falling watermelon (\cref{fig:teaser,fig:beta}), where the internal seed and flesh remain distinctly separate. These results demonstrate the generalizability of our framework across diverse material combinations and fracture patterns.


\paragraph{\textbf{Relighting with Phong Shading}}
Our framework can achieve realistic simulations under different lighting conditions in dynamic scenes. To showcase this capability, we integrate a lighting system, \eg, Blinn-Phong shading model, into our framework. This requires accurate surface normals for each \ac{gs}, which we obtain by utilizing \cref{eq:scale_reg_condensed} to promote kernel densification and surface alignment, thereby enabling effective PCA-based normal computation. As demonstrated in \cref{fig:teaser}, falling pineapple blocks cast dynamic shadows onto one another, while \cref{fig:example2} in \supp showcases an orbiting light source illuminating multiple fruits on the table with evolving shadows and highlights. For further details, please refer to \cref{sec:phong} in the \supp.

\section{Conclusion and Discussion}
\label{sec:conclusion}
In this paper, we introduce \modelname, a novel framework for physically plausible and realistic simulations of dynamic scenes with 3D Gaussian Splatting, including material fracture and behaviors of mixed materials.
Our core contributions include a method for internal structure texture synthesis, an adapted CD-MPM for efficient physics simulation.
This integration allows \modelname to simulate complex events like shattering, deformation, fluid splashing, cutting, and granular collapse with high visual fidelity directly within the \ac{gs} representation, as demonstrated by our diverse qualitative results. The ability to model, simulate, and render dynamic scenes paves the way for more applications involving dynamic and interactive virtual worlds.

\paragraph{Limitations and Future Direction}
To further enhance the applicability and generalization of physical simulation in the \ac{gs} framework, we point out several promising directions for future work.
Firstly, enhancing physical accuracy and versatility could be achieved by incorporating a broader range of constitutive models and exploring simulation techniques that are better suited for specific phenomena like fluids and granular materials.
Secondly, the current physical parameters are manually set; automating this process through inverse rendering or learning-based approaches would significantly reduce tuning efforts and could improve simulation fidelity.
Future research could also focus on scalability for extremely complex scenes, more intricate multi-physics interactions, and effectively integrating learning-based methods for predictive simulation.

\small
\bibliographystyle{ieeenat_fullname}
\bibliography{reference}

\clearpage
\appendix
\setcounter{page}{1}
\renewcommand{\thefigure}{S.\arabic{figure}}
\renewcommand{\thetable}{S.\arabic{table}}
\renewcommand{\theequation}{S.\arabic{equation}}
\maketitlesupplementary

\renewcommand\thefigure{A\arabic{figure}}
\setcounter{figure}{0}
\renewcommand\thetable{A\arabic{table}}
\setcounter{table}{0}
\renewcommand\theequation{A\arabic{equation}}
\setcounter{equation}{0}
\setcounter{footnote}{0}


\section{Material Point Method}
\label{appx:mpm}
\paragraph{Overview}
We use an explicit MPM. Particles (also the 3D Gaussian splats) carry

\begin{equation}
m_p, \; V_p^0, \; \mathbf{X}_p, \; \mathbf{x}_p, \; \mathbf{v}_p, \; \mathbf{F}_p, \; \mathbf{A}_p, \; \mathbf{a}_p, \; \mathbf{\sigma}_p, \; \mathbf{C}_p. 
\label{eq:A1}
\end{equation}
In summary, given the 3D GS of a static scene $\{X_p, A_p, \sigma_p, C_p\}$, we use simulation to dynamize the scene by evolving these Gaussians to produce dynamic Gaussians $\{x_p(t), a_p(t), \sigma_p, C_p\}$. Here, \( \mathbf{X}_p \) is the initial  position, while \( \mathbf{x}_p \) is the current  position that evolves over time with velocity \( \mathbf{v}_p \). Furthermore, \( \mathbf{A}_p \) is the static covariance of the initial Gaussian; the dynamic covariance \( \mathbf{a}_p \) is derived at each step; \( \mathbf{F}_p \) is the deformation gradient used to calculate \( \mathbf{a}_p \); and the opacity \( \sigma_p \) and SH coefficient magnitudes \( \mathbf{C}_p \) are considered time-invariant.

\subsection{The Material Point Method (MPM) Algorithm Steps}

The Material Point Method (MPM) algorithm iteratively transfers data between particles and a background grid. A single time step can be broken down into the following three main stages.

\subsubsection{Particle-to-Grid Transfer (P2G)}
In the first stage, information is transferred from the Lagrangian particles to the nodes of the Eulerian grid. This process, known as rasterization, effectively creates a grid-based snapshot of the continuum's state. For each particle \(p\), its mass \(m_p\) and momentum \(\mathbf{p}_p = m_p \mathbf{v}_p\) are interpolated and added to the surrounding grid nodes \(i\). This is done using interpolation functions \(N_{ip}\) (also known as shape functions), which depend on the particle's position relative to the grid.

The nodal mass \(m_i\) and nodal momentum \(\mathbf{p}_i\) are computed as follows:

\begin{equation}
    m_i = \sum_p m_p N_{ip}
    \label{eq:A2}
\end{equation}
\begin{equation}
    \mathbf{p}_i = \sum_p m_p \mathbf{v}_p N_{ip}.
    \label{eq:A3}
\end{equation}
From the nodal momentum and mass, the initial nodal velocity is found: \(\mathbf{v}_i = \mathbf{p}_i / m_i\), provided \(m_i > 0\).

\subsubsection{Grid Update}
This stage contains the core physics computations, which are performed entirely on the grid. First, forces acting on each grid node are calculated. These forces are typically composed of two parts:
\begin{itemize}[leftmargin=*]
    \item \textbf{Internal forces} \(\mathbf{f}_i^{\text{internal}}\), which arise from the material's stress. These are computed by transferring particle stress information (derived from the deformation gradient \(\mathbf{F}_p\)) back to the grid.
    \item \textbf{External forces} \(\mathbf{f}_i^{\text{external}}\), such as gravity or user-defined interactions.
\end{itemize}
The total force on a node is \(\mathbf{f}_i = \mathbf{f}_i^{\text{internal}} + \mathbf{f}_i^{\text{external}}\).

With the total force, the grid node velocities are updated over the time step \(\Delta t\) using an explicit time integration scheme (e.g., Forward Euler):
\begin{equation}
    \mathbf{v}_i^{n+1} = \mathbf{v}_i^n + \Delta t \frac{\mathbf{f}_i}{m_i}.
    \label{eq:A4}
\end{equation}
Boundary conditions, such as collisions with obstacles, are also enforced on the grid during this stage by modifying the nodal velocities.

\subsubsection{Grid-to-Particle Transfer (G2P)}
Finally, the updated kinematic information is transferred from the grid back to the particles. This stage, often called the "gather" step, updates the Lagrangian particles' state using the newly computed fields on the Eulerian grid, preparing them for the next time step. This process involves updating each particle's velocity, its deformation gradient, and finally its position.

First, the particle's velocity \(\mathbf{v}_p\) is updated by interpolating the new velocities \(\mathbf{v}_i^{n+1}\) from the surrounding grid nodes. This is essentially a weighted average, using the same interpolation functions \(N_{ip}\) as the P2G step:
\begin{equation}
    \mathbf{v}_p^{n+1} = \sum_i \mathbf{v}_i^{n+1} N_{ip}.
    \label{eq:A5}
\end{equation}
This update can be a pure Particle-In-Cell (PIC) update, or it can be combined with the particle's previous velocity in a FLIP (Fluid-Implicit-Particle) scheme to reduce numerical dissipation.

Simultaneously, the particle's deformation gradient \(\mathbf{F}_p\), which tracks the local rotation and strain of the material, must also be updated. This is done by first computing the velocity gradient \(\nabla \mathbf{v}\) at the particle's position, which is also interpolated from the grid node velocities:
\begin{equation}
    \nabla \mathbf{v}_p = \sum_i \mathbf{v}_i^{n+1} \nabla N_{ip}^T.
\end{equation}
This gradient is then used to advance the deformation gradient forward in time:
\begin{equation}
    \mathbf{F}_p^{n+1} = \left( \mathbf{I} + \Delta t \, \nabla \mathbf{v}_p \right) \mathbf{F}_p^n,
\end{equation}
where \(\mathbf{I}\) is the identity matrix. This update is crucial for correctly computing material stress in the next time step.

Lastly, with the new velocity \(\mathbf{v}_p^{n+1}\) computed, the particle's position \(\mathbf{x}_p\) is updated as:
\begin{equation}
    \mathbf{x}_p^{n+1} = \mathbf{x}_p^n + \Delta t \, \mathbf{v}_p^{n+1}.
\end{equation}
Once all particles have been updated, the information on the background grid is no longer needed and is typically reset or discarded. The simulation is now ready to begin the next time step with the P2G phase.

\subsection{Evolution of 3D Gaussian Properties via Continuum Mechanics}

This approach outlines a method for animating 3D GS by treating them as discrete particles within a physics-based system governed by continuum mechanics. The primary goal is to evolve a static scene, defined by initial properties, into a dynamic state for rendering.

The evolution of the key Gaussian properties for each time step is as follows:

\begin{itemize}[leftmargin=*]
    \item \textbf{Position Evolution (Mean):} The Gaussian's center, or mean, is its world-space position \(\mathbf{x}_p\). This is updated using the particle's velocity \(\mathbf{v}_p\), which is determined by the physical simulation, via explicit time integration:
    \begin{equation}
        \mathbf{x}_p^{n+1} = \mathbf{x}_p^n + \Delta t \, \mathbf{v}_p.
    \end{equation}

    \item \textbf{Shape Evolution (Covariance):} The dynamic world-space covariance \(\mathbf{a}_p\), which defines the Gaussian's shape and size, is computed directly from the deformation gradient \(\mathbf{F}_p\). The deformation gradient describes the local deformation of the material around the particle. It maps the initial, undeformed shape (defined by the material-space covariance \(\mathbf{A}_p\)) to its current, deformed configuration:
    \begin{equation}
        \mathbf{a}_p(t) = \mathbf{F}_p(t) \mathbf{A}_p \mathbf{F}_p(t)^T.
    \end{equation}

    \item \textbf{Orientation Evolution (for Rendering):} To correctly render anisotropic appearances (e.g., using Spherical Harmonics), the particle's orientation must be tracked. The rotation component \(\mathbf{R}_p\) is extracted from the deformation gradient, typically via polar decomposition (\(\mathbf{F}_p = \mathbf{R}_p \mathbf{S}_p\)). This rotation is then applied to the appearance model during rendering.

    \item \textbf{Time-Invariant Properties:} Visual attributes such as opacity \(\sigma_p\) and material-space appearance coefficients (e.g., Spherical Harmonics, \(\mathbf{C}_p\)) are considered intrinsic material properties. They are typically held constant throughout the simulation.
\end{itemize}

\section{Fracture mechanism with \texorpdfstring{\acl{cdmpm}}{CD-MPM}}
\label{sec:cdmpm_supp}

\subsection{Introduction of CD-MPM}
The yield surface serves as a dividing boundary in stress space: inside it, the material response is elastic; at the boundary plastic yielding begins; any trial state predicted beyond this boundary is reconciled by returning it to a suitable point on the boundary in accordance with ideal plasticity. As mentioned above, the yield surface of CD‑MPM is defined as:
\begin{equation}
  y(p,q) = (1+2\beta)\,q^{2} + M^{2}(p+\beta p_{0})(p-p_{0})= 0.
\end{equation}
 If \((p,q)\) lies in the elastic domain where $y\le0$, no plastic correction is applied.

%
\begin{equation}
  (p_{c}, q_{c}) = \Big(\frac{1-\beta}{2}p_{0},\,0\Big)
\end{equation}
%
\begin{equation}
 y_{tr} = y(p_{tr}, q_{tr})
\end{equation}
%
\begin{equation}
  J_{E}(p) = \sqrt{-\frac{2p}{\kappa} + 1}
\end{equation}

Here \(p_c, q_c\) identify the center of the yield ellipsoid (\(y=0\)); \(p_{tr}, q_{tr}\) is the uncorrected trial stress state produced at simulation step \(n\); \(J_E\) is the determinant of the elastic deformation gradient (elastic volume ratio); \(\kappa\) is the  Bulk Modules; and \(p_{n+1}, q_{n+1}\) is the  state after applying the return mapping \(R\):

\begin{equation}
R(p_{n+1}, q_{n+1}) =
\left\{
\begin{array}{l@{\quad}l}
(p_{tr}, q_{tr}), & y_{tr} \le 0 \\[2pt]
& \text{(Elastic)} \\[4pt]
(p_0, 0), & y_{tr} > 0 \ \wedge \ p_{tr} > p_0 \\[2pt]
& \begin{array}{@{}l@{}} \text{(Case 1: upper tip} \\ \text{\phantom{(Case 1: }projection)} \end{array} \\[4pt]
(-\beta p_0, 0), & y_{tr} > 0 \ \wedge \ p_{tr} < -\beta p_0 \\[2pt]
& \begin{array}{@{}l@{}} \text{(Case 2: lower tip} \\ \text{\phantom{(Case 2: }projection)} \end{array} \\[4pt]
(p_x, q_x), & y_{tr} > 0 \ \wedge \ -\beta p_0 \le p_{tr} \le p_0 \\[2pt]
& \begin{array}{@{}l@{}} \text{(Case 3: center--trial line} \\ \text{\phantom{(Case 3: }intersection)} \end{array}
\end{array}
\right.
\end{equation}

Here \(y_{tr}=y(p_{tr}, q_{tr})\). If \(y_{tr}\le 0\), the trial point lies in the elastic domain and is accepted unchanged: \((p_{n+1}, q_{n+1})=(p_{tr}, q_{tr})\). If \(y_{tr}>0\) and \(p_{tr}>p_{0}\), the trial point lies beyond the positive \(p\)-axis tip and is projected to the upper tip \((p_{0},0)\). If \(y_{tr}>0\) and \(p_{tr}< -\beta p_{0}\), it lies beyond the negative tip and is projected to \((-\beta p_{0},0)\). Otherwise (\(y_{tr}>0\) with \(-\beta p_{0} \le p_{tr} \le p_{0}\)), we join the center \((p_{c}, q_{c})\) and the trial point \((p_{tr}, q_{tr})\); the intersection of this line segment with the yield ellipsoid \(y(p,q)=0\) defines \((p_{x}, q_{x})\), and we set \((p_{n+1}, q_{n+1})=(p_{x}, q_{x})\). Besides \(p, q\), we also update \(\alpha\) and \(J_E\) as below:

\begin{equation}
  \alpha_{n+1} =
  \alpha_{n} +
  \begin{cases}
    0, & y_{tr} \le 0 \\[4pt]
    \log\!\big(J_{E,tr} / J_{E,n+1}\big), & y_{tr} > 0
  \end{cases},
\end{equation}
with
\begin{equation}
  J_{E,n+1} =
    \begin{cases}
      J_{E}(p_{0}), & \text{Case 1} \\
      J_{E}(-\beta p_{0}), & \text{Case 2} \\
      J_{E}(p_{x}), & \text{Case 3}
    \end{cases}
\end{equation}

\subsection{Adapted Continuous Return Mapping}
However, this piecewise return mapping is discontinuous at the right tip \(p=p_{0}\). Consider trial states with \(y_{tr}>0\) and very large shear measure \(q_{tr}\to\infty\). Take two sequences with \(p_{tr}=p_{0}-\varepsilon\) and \(p_{tr}=p_{0}+\varepsilon\) (\(\varepsilon>0\)). For \(p_{tr}=p_{0}-\varepsilon\), the algorithm falls into the “center–trial line intersection” branch; as \(q_{tr}\to\infty\) the direction from the center \((p_{c},0)\), with \(p_{c}=\tfrac{1-\beta}{2}p_{0}\), to the trial point becomes vertical, so the mapped point tends to the upper apex of the yield ellipsoid \(y(p,q)=0\), namely,
\begin{equation}
\begin{split}
\lim_{\varepsilon \to 0^{+}} \lim_{q_{tr} \to \infty} R(p_{0} - \varepsilon, q_{tr}) &=
\Big( \tfrac{1-\beta}{2} p_{0}, \ \tfrac{M(\beta+1)}{2\sqrt{1+2\beta}} \, p_{0} \Big), \\
\lim_{\varepsilon \to 0^{+}} \lim_{q_{tr} \to \infty} R(p_{0} + \varepsilon, q_{tr}) &= (p_{0}, 0),
\end{split}
\end{equation}

\begin{table*}[t] 
\centering

\caption{\textbf{Parameters and Timings.} Seconds per frame (s/frame) is an average. All performance metrics were obtained from experiments conducted on a GPU delivering 103 Tensor TFLOPS at FP16 precision.}
\resizebox{\linewidth}{!}{
\begin{tabular}{@{}l l c c c c c c c c@{}} 
\toprule
Example & s/frame & $\Delta t_{frame}$ & $\Delta x$ & $\Delta t_{step}$ & $N$ & $\rho$ & $E$ & $\nu$ & NACC-($\alpha_0$, $\beta$, $\xi$, $M$) \\
\midrule
watermelon & 3.56 & 1/50 & $3 \times 10^{-3}$ & $1 \times 10^{-4}$ & 27M & 2 & 2000/1000/$1 \times 10^4$ & 0.38 & (-0.04, 2/0.6/5, 2, 2.36) \\
jelly & 0.39 & 1/500 & $3 \times 10^{-3}$ & $1 \times 10^{-5}$ & 1M & 2 & 2000 & 0.45 & (-0.5, 1, 2, 2.36) \\
pumpkin & 5.12 & 1/50 & $3 \times 10^{-3}$ & $1 \times 10^{-4}$ & 27M & 2 & 4000 & 0.40 & (-0.04, 1, 2, 2.36) \\
kiwi & 1.58 & 1/50 & $1 \times 10^{-2}$ & $1 \times 10^{-4}$ & 1M & 2 & 2000 & 0.42 & (-0.04, 1, 2, 2.36) \\
pineapple & 1.16 & 1/50 & $1 \times 10^{-2}$ & $1 \times 10^{-4}$ & 1M & 2 & 5000 & 0.39 & (-0.04, 1, 2, 2.36) \\
dragonfruit & 2.27 & 1/50 & $1 \times 10^{-2}$ & $1 \times 10^{-4}$ & 1M & 2 & 2000 & 0.42 & (-0.04, 1, 2, 2.36) \\
tosta & 3.18 & 1/50 & $5 \times 10^{-3}$ & $1 \times 10^{-4}$ & 8M & 2 & 2000 & 0.38 & (-0.1, 1, 2, 2.36) \\
sandcastle & 2.09 & 1/50 & $1 \times 10^{-2}$ & $1 \times 10^{-4}$ & 8M & 2 & 50 & 0.05 & (-0.04, 0.01, 1, 2.36) \\
\bottomrule
\end{tabular}}
\label{tab:parameters}
\end{table*}
showing a directional jump: one limit preserves (essentially) shear while the other preserves only the volumetric extension. And even some small \(q\) such as \(q=p_0\) will also occur jumps like this. 

\begin{figure}[htbp]
\centering 
    \includegraphics[width=\linewidth]{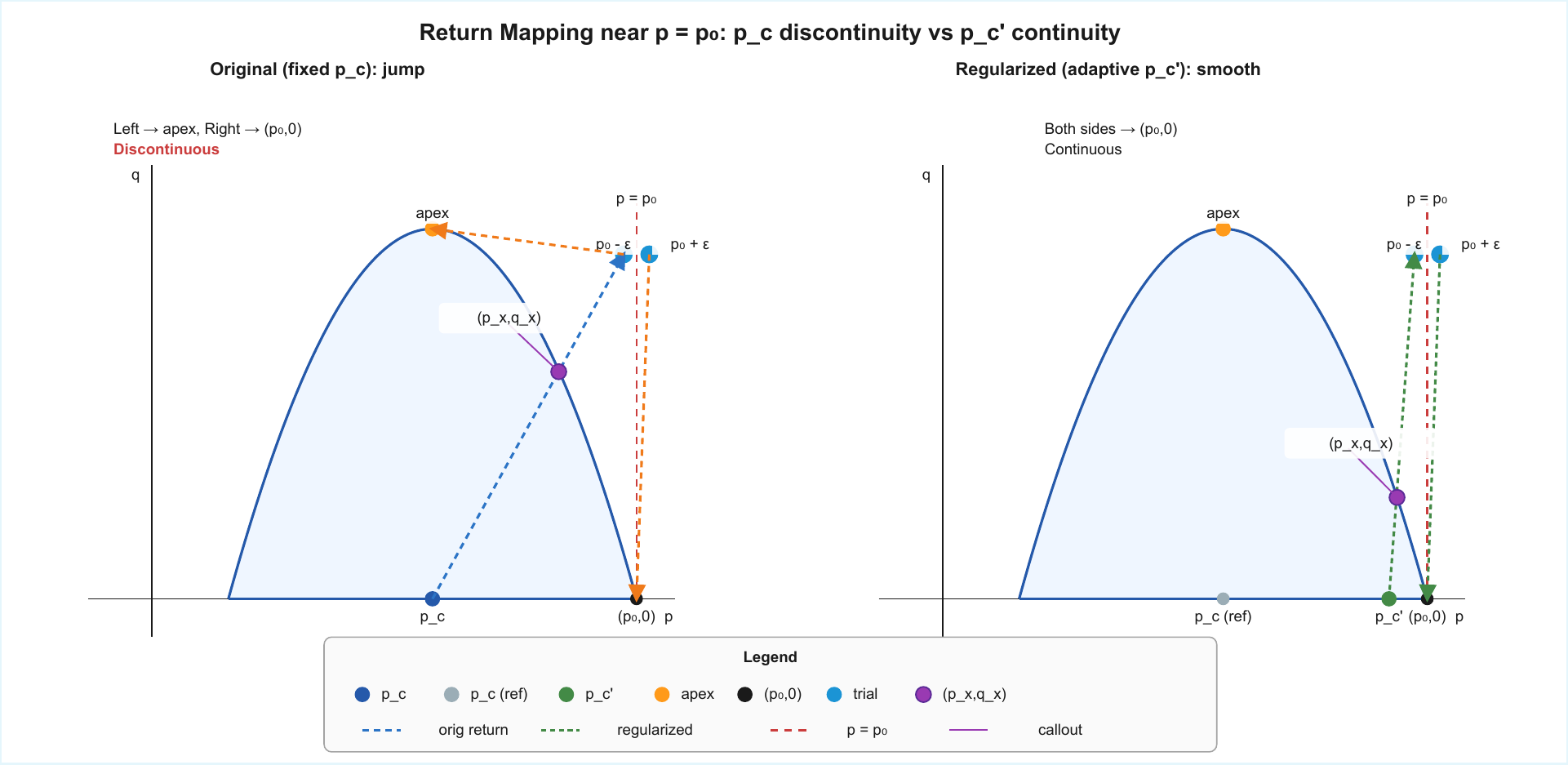}
    \caption{Comparison of two return mapping kinds.}
    \label{fig:return_mapping}
\end{figure}

To remove both the numerical instability and the physical ambiguity at the tip, we replace the interior (\(p_{tr}\in[-\beta p_{0},p_{0}]\)) center–line branch with a normal closest-point return: solve \( (p_{n+1},q_{n+1})=(p_{tr},q_{tr})-\Delta\lambda\nabla y(p_{n+1},q_{n+1}) \), \(y(p_{n+1},q_{n+1})=0\), \(\Delta\lambda\ge 0\). Outside this interval, we still project to the nearest tip. This yields a continuous mapping and a well-defined consistent tangent.

We modify only the interior plastic branch with $-\beta p_{0} \le p_{tr} \le p_{0}$.
Introduce a \(k\)-dependent pseudo-center on the \(p\)-axis:
\begin{equation}
\begin{gathered}
L_p = p_{0} - p_{c} > 0, \\
\phi_{k} = \left| \frac{p_{tr} - p_{c}}{L_p} \right|^{k} \in [0,1], \\
(p_c', q_c') = \big(p_{c} + \phi_{k} (p_{tr} - p_{c}), \, 0 \big).
\end{gathered}
\end{equation}

In Case 3 we replace the fixed center \((p_c,0)\) by \((p_c',0)\), draw the line through \((p_c',0)\) and the trial point \((p_{tr}, q_{tr})\), and take its intersection with the yield surface \(y=0\) as the updated stress, as shown in \cref{fig:return_mapping}. All other cases are unchanged. For any finite \(k\) the return mapping is continuous, because as \(p_{tr}\to p_{0}^{-}\) we have \(\phi_k \to 1\) and thus \(p_c' \to p_{tr}\), so the update approaches the right tip smoothly. For any fixed interior \(p_{tr}<p_{0}\), \(\phi_k \to 0\) as \(k\to\infty\), giving \(p_c' \to p_c\) and recovering the original (unmodified) branch. Hence \(k\) provides a homotopy from a continuous regularized mapping (finite \(k\)) back to the original formulation (\(k\to\infty\)).

\subsection{GPU Parallelization} 
\label{appx:gpu_parallel}
We achieve a substantial performance improvement by porting the  CPU-bound CD-MPM algorithm to the GPU. \textbf{Our implementation reduces simulation times from 4 minutes per frame to a single second}. This is accomplished through a complete framework reimplementation that leverages the NVIDIA Warp library to parallelize the core simulation loop. Unlike the original CPU-only method, our GPU-native approach enables the simulation of far more complex scenes in interactive time.

\section{Experiment details }
\label{appx:details}
All experiments are conducted on a  GPU capable of 52.22 TFLOPS (FP32) and approximately 103 Tensor TFLOPS (FP16). These simulations typically consume around 10 GB of VRAM, with peak usage not exceeding 16 GB. Detailed timings and material parameters are provided in ~\cref{tab:parameters}. For the \ac{nacc} model, the parameter $\beta$ is adjusted to differentiate the material properties of various components, while the initial parameter $\alpha_0$ is maintained uniformly for all particles within an object.

For coarse texture generation, MVInpainter~\citep{shi2023mvdream} is selected over IP-Adapter~\citep{ye2023ip} and MVDream~\citep{shi2023mvdream} due to its ability to maintain color consistency across different viewing axes. Subsequently, SD-XL was employed for fine texture generation, owing to its enhanced performance in generating detailed interior textures compared to IP-Adapter.

For the user study, we prepare eight distinct objects: watermelon, cake, jelly, pumpkin, bread, kiwi, dragonfruit, and pineapple. 
We then conduct two separate evaluations. 
To assess the quality of the interior filling, we recruit 21 participants, collecting a total of $8 \times 21 = 168$ ratings. 
Separately, to evaluate the simulation dynamics, 26 participants are recruited, providing a total of $8 \times 26 = 208$ ratings.

We use two types of prompts:
\begin{itemize}[leftmargin=*]
\item \textbf{For interior filling}: We explicitly instruct GPT to generate an inpainting prompt in the form “a slice of [object].” For example, GPT produces the following for a watermelon: “A realistic and detailed drawing of the juicy red flesh and black seeds of a watermelon slice.”

\item \textbf{For CLIP-score evaluation}: To evaluate the plausibility of the final scene, we have human annotators write prompts that describe the overall event, for example, “A watermelon dropped and shattered on a table,” and “Slices of a [object] landing on a table.”
\end{itemize}

\subsection{Relighting}
\label{sec:phong}
Existing \ac{gs} lighting methods, such as Relightable \ac{3dgs}~\citep{gao2024relightable} and GS-Phong~\citep{he2024gsphong}, are not applicable to dynamic simulations. These methods are intended for static scenes and rely on multiple images captured under known lighting conditions to learn \ac{gs} normals and other attributes. Rather than adopting the \ac{pbr} lighting model in Relightable \ac{3dgs}, which requires learning additional material attributes, \eg, Fresnel parameters, for each \ac{gs}, we employ the empirical \phong, which only requires the normals for \ac{gs}.

However, it is nontrivial to obtain \ac{gs} normals using non-learning methods. As noted in Relightable \ac{3dgs}, numerical normal-estimation methods such as PCA are ill-suited to \ac{gs} for two primary reasons: (i) \ac{gs} particles are spatially sparse, and (ii) Gaussian centers, especially those with large kernels, are not tightly aligned with the visual surface. To overcome these issues, the regularization loss~\ref{eq:scale_reg_condensed} we introduce in \cref{sec:gs_inpaint} promotes kernel densification and surface alignment, thereby enabling effective normal computation for each Gaussian splat using PCA.

\paragraph{Blinn-Phong Reflection Model}
Once the normal $\mathbf n$ for each \ac{gs} is computed, we apply the \phong to determine its final color.  For each Gaussian $i$ with center $\mathbf p_i$ and normal $\mathbf n_i$, we apply the \phong  using view direction $\mathbf v$ (from $\mathbf p_i$ to the camera) and, for each light $m$, light direction $\mathbf l_m$, distance $r_m$, and half vector
$\mathbf h_m = (\mathbf l_m + \mathbf v)/\lVert \mathbf l_m + \mathbf v \rVert_2$.
The diffuse and specular terms are
$D_m = \max(\mathbf n_i \cdot \mathbf l_m, 0)$ and
$S_m = [\max(\mathbf n_i \cdot \mathbf h_m, 0)]^{p}$, with shininess exponent $p$.
Let $\mathbf c_0$ be the base color, $\mathbf I_a$ the ambient light color,
$\mathbf I_{L,m}$ the color of light $m$, and $T_{i,m}$ a per-light visibility term.
Then
\begin{equation}
\mathbf L_i = \mathbf c_0 \odot \mathbf I_a
+ \sum_m T_{i,m}\,(\mathbf c_0 \odot \mathbf I_{L,m}) \frac{1}{r_m^{2}}\,( D_m + S_m ),
\label{eq:blinn_phong_compact}
\end{equation}
where $\odot$ denotes element-wise multiplication.

\begin{figure*}[htbp]
\centering 
    \includegraphics[width=\textwidth]{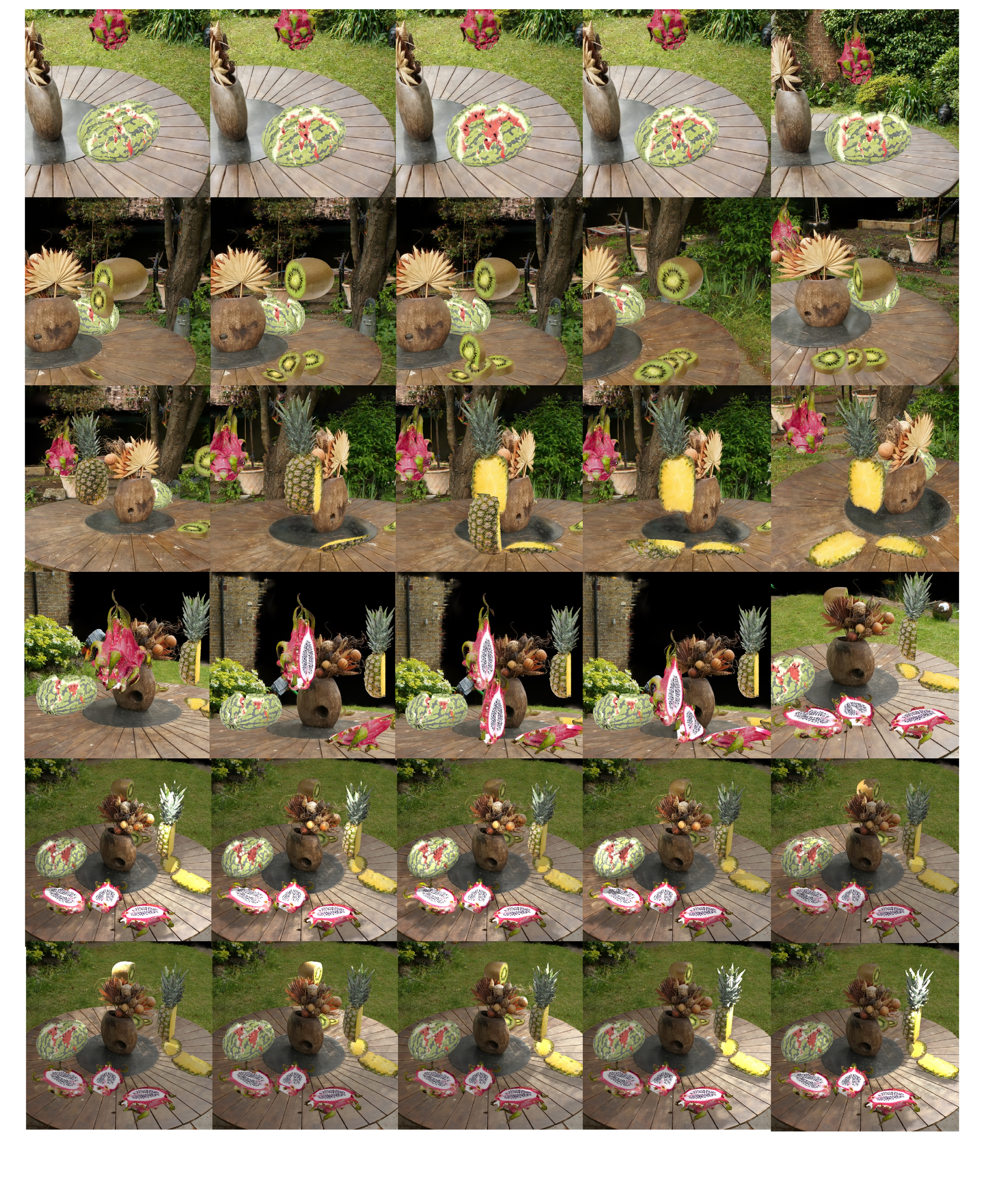}
    \caption{More examples of object simulation and illumination.}
    \label{fig:example2}
\end{figure*}

This lighting framework allows us to effectively simulate complex scenes with multiple objects and dynamic light sources, as shown in ~\cref{fig:teaser} and ~\cref{fig:example2}. For example, in the latter figure, we present a scene of multiple fruits with dynamic lighting on a table. Such dynamic illumination and shadowing are crucial for achieving visually consistent and plausible renderings during simulation, where the evolution of shadows is not considered in PhysGaussian~\citep{xie2024physgaussian}.

\section{Use of Large Language Models (LLMs)}
We used a large language model solely as a writing aid to improve the clarity, grammar, and overall readability of the manuscript. Its role was limited to polishing the language and refining sentence structure, without contributing to research ideation, experimental design, or data analysis. All technical ideas, methods, results, and conclusions are entirely the work of the authors, and we take full responsibility for the final content.

\clearpage
\begin{figure*}[htbp]
\centering 
    \includegraphics[width=\textwidth]{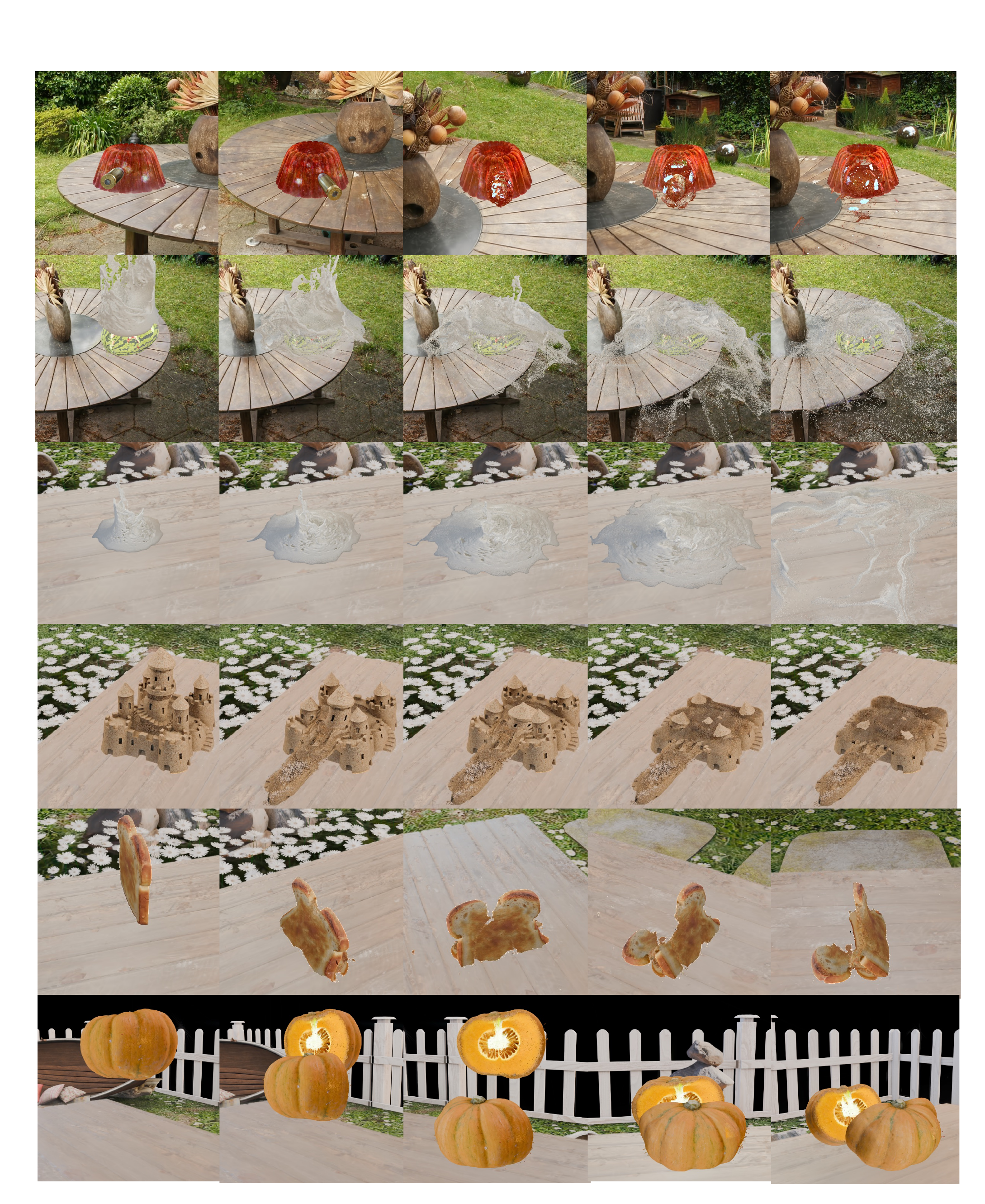}
    \caption{More examples of object simulation.}
    \label{fig:example}
\end{figure*}

\end{document}